\documentclass{article}

\usepackage[final]{neurips_data_2023}

\usepackage[utf8]{inputenc} 
\usepackage[T1]{fontenc}    
\usepackage{hyperref}       
\usepackage{url}            
\usepackage{booktabs}       
\usepackage{amsfonts}       
\usepackage{nicefrac}       
\usepackage{microtype}      
\usepackage{xcolor}         

\usepackage[style=ieee]{biblatex}
\usepackage{bibentry}
\usepackage{graphicx}
\usepackage{subfig}

\usepackage{array}
\newcolumntype{L}[1]{>{\raggedright\arraybackslash}m{#1}}

\title{InDL: A New Dataset and Benchmark for In-Diagram Logic Interpretation based on Visual Illusion}

\author{%
  Haobo Yang \\
  School of Informatics \\
  The University of Edinburgh \\
  \texttt{s1911593@ed.ac.uk}
  \And
  Wenyu Wang \\
  School of Philosophy, \\ Psychology 
  and Language Sciences \\
  The University of Edinburgh \\
  \texttt{s2103736@ed.ac.uk } \\
  \And
  Ze Cao \\
  School of Informatics \\
  The University of Edinburgh \\
  \texttt{s1973433@ed.ac.uk} \\
  \And
  Zhekai Duan \\
  School of Enginnering \\
  The University of Edinburgh \\
  \texttt{s2085313@ed.ac.uk } \\
  \And
  Xuchen Liu \\
  School of Informatics \\
  The University of Edinburgh \\
  \texttt{s2420193@ed.ac.uk } \\
}

\begin{document}

\maketitle

\section{Abstract}

This paper introduces a novel approach to evaluating deep learning models' capacity for in-diagram logic interpretation. Leveraging the intriguing realm of visual illusions, we establish a unique dataset, InDL, designed to rigorously test and benchmark these models. Deep learning has witnessed remarkable progress in domains such as computer vision and natural language processing. However, models often stumble in tasks requiring logical reasoning due to their inherent 'black box' characteristics, which obscure the decision-making process. Our work presents a new lens to understand these models better by focusing on their handling of visual illusions -- a complex interplay of perception and logic. We utilize six classic geometric optical illusions to create a comparative framework between human and machine visual perception. This methodology offers a quantifiable measure to rank models, elucidating potential weaknesses and providing actionable insights for model improvements. Our experimental results affirm the efficacy of our benchmarking strategy, demonstrating its ability to effectively rank models based on their logic interpretation ability. As part of our commitment to reproducible research, the source code and datasets will be made publicly available here: \href{https://github.com/rabbit-magic-wh/InDL}{https://github.com/rabbit-magic-wh/InDL}.

\section{Introduction}
Deep learning, a subfield of artificial intelligence, has demonstrated impressive capabilities in solving intricate problems across various domains such as computer vision and natural language processing. Despite these advancements, the inner workings of deep learning models often remain obscured, leading to what is commonly referred to as the 'black box' dilemma. This lack of transparency in decision-making processes is particularly evident when logical reasoning is required, a limitation that grows more pressing as the complexity of tasks assigned to these models increases.

In response to these challenges, we introduce a novel research approach grounded in principles borrowed from psychology. Our methodology challenges deep learning models to grapple with in-diagram logic interpretation through a series of visual illusions, offering insights into the models' understanding of logical reasoning within the context of visual perception. We posit that, irrespective of their 'black box' nature, deep learning models should adhere to logical principles, at least in terms of their output.

Furthermore, we seek to explore the extent to which phenomena observed in humans, such as the perception of visual illusions, are replicated within deep learning models. This investigation allows us to draw parallels between human cognition and artificial intelligence, fostering a richer understanding of how these systems perceive and interpret information.

Our contributions are two-fold. Firstly, we propose a unique methodology for assessing the logical reasoning capabilities of deep learning models, casting light on their otherwise opaque decision-making processes. Secondly, we establish a systematic framework for quantifying the logical comprehension of these models. Until now, the diversity and complexity of input datasets have made it challenging to measure these capabilities quantitatively, often leading to discussions of results without a solid analysis of input-output relationships.

Our methodology allows for a comprehensive analysis of both inputs (causes) and outputs (effects), thus providing a more quantitative perspective on a model's understanding of logic. Through rigorous experiments, we demonstrate the efficacy of our proposed framework, generating a ranking of models based on their abilities to interpret in-diagram logic. This not only uncovers potential weaknesses within these models but also paves the way for potential improvements.

The remainder of this paper unfolds as follows: we begin with a review of related work in the field of deep learning and logic interpretation, followed by a detailed presentation of our proposed evaluation methodology. Next, we delve into our experimental design, discussing the results obtained, and conclude by presenting our findings and outlining directions for future work.
\section{Related Works}

\subsection{History of Logic Interpretation}
The field of machine learning and neural networks has observed a series of advancements and recessions since its inception in the 1950s. Early neural network models, such as the Perceptron \cite{rosenblatt1958perceptron}, demonstrated commendable proficiency in tackling linear classification problems but failed to address more complex non-linear problems like the XOR logic function \cite{minsky69perceptrons}. As the exploration into neural networks deepened, techniques like Backpropagation(BP) and Multi-layer Perceptron (MLP) \cite{rumelhart1986learning} surfaced, bringing potential solutions to non-linear problems. However, a myriad of challenges persist in the application of neural networks to logical problems, including understanding logical structures, symbolic reasoning, generalisation capabilities, interpretability, explainability, training data bias, and computational resource constraints.

A variety of methods have been proposed by the academic community to overcome these challenges. In understanding logical structures, researchers have explored the use of Graph Neural Networks (GNNs) \cite{Scarselli:2009ku} and Recursive Neural Networks (RNNs) \cite{10.5555/3104482.3104499} to represent and process hierarchical logical relationships. For symbolic reasoning, neuro-symbolic integration methods have been suggested, fusing neural networks with symbol-based logical reasoning systems to harness the advantages of both \cite{garnelo2016deep}\cite{besold2017neuralsymbolic}). Meta-Learning \cite{pmlr-v70-finn17a} and Transfer Learning \cite{PanY09TKDE} techniques have been utilized to enhance generalization capabilities, enabling models to better manage novel and intricate logical problems.

Assessing the efficacy of neural networks in addressing logical problems primarily involves metrics such as accuracy, generalisation capabilities, and execution time. To facilitate a comprehensive evaluation of model performance, researchers have designed a multitude of benchmark datasets, including CLEVR\cite{johnson2016clevr} and NLVR \cite{suhr-etal-2017-corpus}, which focus on visual reasoning and natural language reasoning tasks respectively. These benchmark datasets encompass a wide array of logical problems, presenting various levels of difficulty and complexity, thereby enabling researchers to contrast and assess the effectiveness of different methods.

\subsection{The Black Box Problem in Deep Learning for Logic}
Despite the significant advancements made by deep learning models in areas such as computer vision \cite{lecun2015deep} and natural language processing \cite{NIPS2017_3f5ee243} over the past few decades, they still face considerable challenges when applied to logical problems. The black box issue of deep learning models complicates the tracking and analysis of the reasoning process in logical problems, making it difficult to ensure consistent adherence to logical rules \cite{Castelvecchi2016CanWO}.

Existing interpretability methods, such as LIME \cite{ribeiro2016why} and SHAP \cite{10.5555/3295222.3295230}, aim to tackle the explainability issue inherent in deep learning models. However, they exhibit limitations in providing comprehensive logical reasoning explanations. Furthermore, the reasoning capabilities of the models might be constrained by the patterns learned from data, reflecting the influence of data distribution and potential bias \cite{conf/cvpr/TorralbaE11}.

These limitations become particularly pronounced when dealing with rigorous logical reasoning problems. For example, in analogy reasoning tasks, deep learning models may be distracted by superficial features of the training data, leading to inaccurate recognition of underlying logical relationships \cite{barrett2018measuring}. Conversely, when dealing with less common or uncharacteristic logical problems, the models might struggle to generalise to new problems \cite{geirhos2022imagenettrained}.

To address the black box problem in logical problems, it is critical to leverage psychologically inspired datasets to ascertain if models demonstrate capabilities in recognising logical tasks that are on par with humans or state-of-the-art (Sota) methods. This approach is expected to drive further advancement in the application of deep learning to logical reasoning.

\subsection{Psychology Background}

While neural networks and deep learning models excel at computer vision tasks, they have a tendency to overlook the underlying logic of images in the psychology area. These models sit at the intersection of neuroscience and psychology, allowing researchers to test hypotheses and predict real-world outcomes through computer simulations \cite{nematzadeh_powers_2020}. In neuropsychology vision study, the primary visual cortex is a crucial biological structure that plays a critical role in a variety of visual tasks, such as object recognition, contextual modulation, and luminance perception. Neuroscience research has extensively explored unsupervised learning within the primary visual cortex, including the emergence of visual illusions. This is evidenced by six classic geometric visual illusions.

Six classic geometric visual illusions \cite{mazumdar_mitra_mandal_ghosh_bhaumik_2022b} are as follows : (A) The Hering illusion \cite{coren1970lateral}. (B) The Wundt illusion \cite{coren1970lateral}. (C) The Muller-Lyer illusion.  (D) The Poggendorff illusion \cite{zanuttini_1976}. (E) The Vertical-horizontal illusion \cite{kunnapas1955analysis}. (F) The Zollner illusion \cite{wallace1965measurements}. The psychological explanations of these geometric optical illusions (GOIs) are mainly based on neuro-mathematical models. These illusions occur due to a mismatch between the geometric properties of a visual stimulus and its associated perception. Moreover, the environment surrounding the visual stimulus can also modify visual perception. Experimental observations have shown that the context and surrounding distractors can modify both visual perception and primary visual cortical responses \cite{akiyama_yamamoto_amano_taketomi_plopski_christian_kato_2018}.

This paper does not delve into the underlying mechanisms, but rather only compares the experimental findings of psychology and AI to compile a list. Though not discussed in this paper, it is relevant for further discussion. A team led by Serre at the Brancani Institute for Brain Science developed a computational model in 2018 that is constrained by data on visual cortical anatomy and neurophysiology. The model aims to capture how neighboring cortical neurons communicate and adjust their responses to each other in response to complex stimuli like contextual visual illusions \cite{NIPS2017_3f5ee243}. However, recent research has shown that there is still a significant cognitive gap between artificial intelligence and humans and that deep neural networks do not exhibit human-like phenomena for illusion contours \cite{gomez-villa_martín_vazquez-corral_bertalmío_2019}. This study proposes a method to transform machine-learning visual datasets into illusion contour samples, inspired by the widespread occurrence of illusion contours in human and biological visual systems. The study quantitatively measures the ability of current deep learning models to recognize illusion contours. The results of the experiments demonstrate that from classical to state-of-the-art deep neural networks, machines are far from as effective as humans in recognizing illusion contours. Therefore, this article aims to compile a list of studies that draw on psychological research methods to develop a study comparing the differences between the human and computer vision systems in six geometric illusions.

\section{Benchmark Specification}

\subsection{Dataset Design}

\begin{table}[!ht]
\centering
\caption{Summary of the five optical illusion datasets used in the study. Each dataset corresponds to a different type of optical illusion, characterized by its unique features. The independent variable represents the primary element manipulated in each experiment, while the controlled variable signifies the aspects that were kept constant. This table provides a clear overview of the experimental design and the varying parameters for each optical illusion.}
\label{tab:dataset_summary}
\resizebox{\linewidth}{!}{
\begin{tabular}{L{0.015\columnwidth}L{0.15\columnwidth}L{0.3\columnwidth}L{0.25\columnwidth}L{0.25\columnwidth}}
\hline
 & \textbf{Dataset} & \textbf{Description} & \textbf{Independent Variable} & \textbf{Controlled Variable} \\ \hline
01 & Hering \& Wundt Illusion & Two parallel lines divided by many lines intersecting in the middle & Angle and density of intersecting lines & Distance and length of parallel lines \\ \hline
02 & Muller-Lyer Illusion & Two parallel lines of identical length, featuring arrows pointing inward and outward at their ends & Angle of arrows & Length of two parallel lines \\ \hline
03 & Poggendorff Illusion & A set of interrupted oblique lines & Angle of oblique line & Distance between parallel lines \\ \hline
04 & Vertical-horizontal Illusion & An L-shaped stimulus created by juxtaposing a horizontal line and a vertical line of equal lengths & Intersection point of horizontal and vertical line & Length of horizontal and vertical line \\ \hline
05 & Zollner Illusion & A set of black, uniformly straight lines, manipulated in stimulus configuration & Angle of stimuli relative to the vertical direction & Position of intersection point \\ \hline
\end{tabular}
}
\end{table}

\begin{figure}[!ht]
  \centering
  \includegraphics[width=0.85\columnwidth]{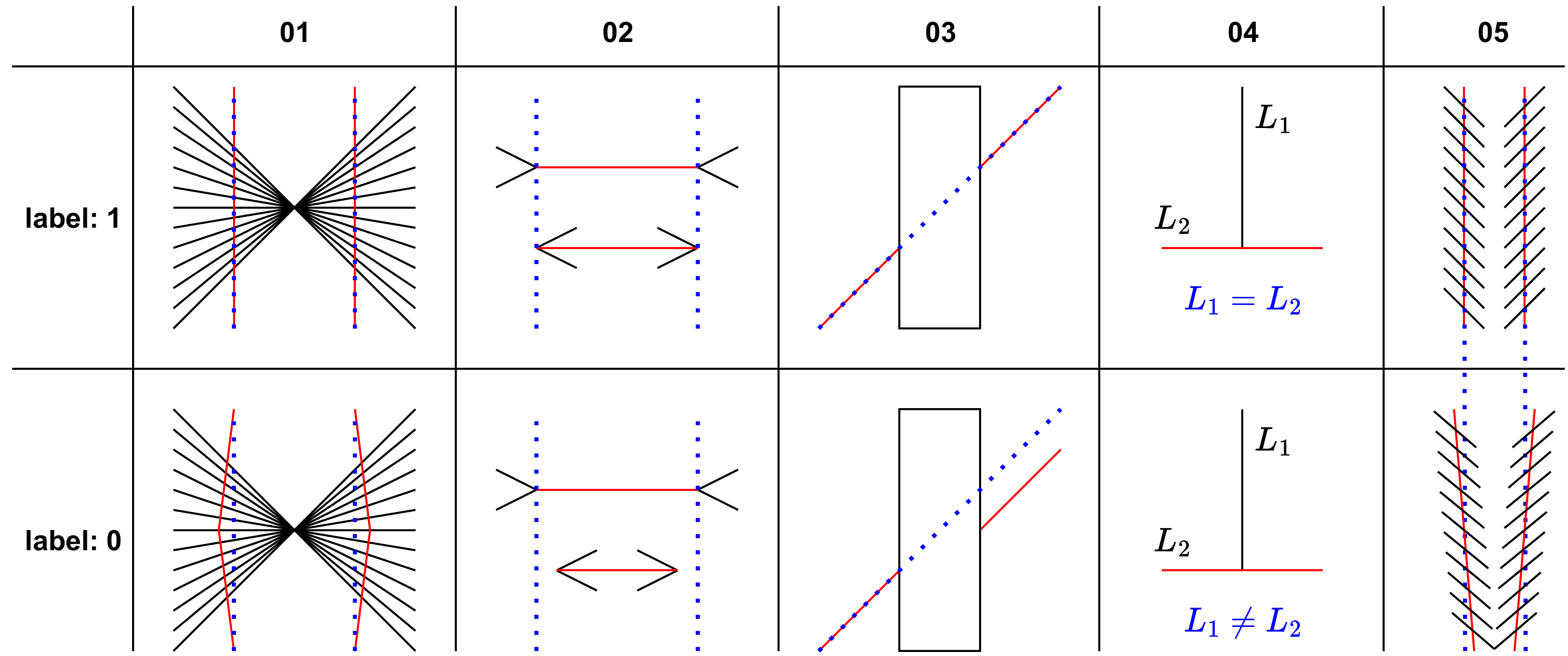}
  \caption{The 5 dataset images, with corresponding labels.}
  \label{distribution-figure}
\end{figure}

Table~\ref{tab:dataset_summary} provides a summary of the five different datasets employed in this study, each associated with a distinct optical illusion, namely the Hering \& Wundt illusion, the Muller-Lyer illusion, the Poggendorff illusion, the Vertical-horizontal illusion, and the Zollner illusion. These illusions were selected due to their unique and characteristic manipulations of visual perception.

The Hering \& Wundt illusion dataset involves two parallel lines divided by several intersecting lines in the middle. The independent variable manipulated in this illusion is the angle and density of the intersecting lines, while the distance and length of the parallel lines are kept constant \cite{coren1970lateral,smeets2004curved,holt-hansen_1961}.

The Muller-Lyer illusion dataset comprises two parallel lines of identical length, terminated with inward and outward arrows at their respective ends. The independent variable is the angle of the arrows, and the length of the two parallel lines is controlled \cite{mazumdar_mitra_mandal_ghosh_bhaumik_2022,coren1970lateral}.

The Poggendorff illusion dataset is based on an interrupted oblique line. The angle of the oblique line is manipulated as the independent variable, while the distance between the parallel lines is held constant \cite{zanuttini_1976,weintraub1980poggendorff,gillam1971depth,howe_yang_purves_2005}.

The Vertical–horizontal illusion dataset employs an L-shaped stimulus created by juxtaposing a horizontal line and a vertical line of equal lengths. The point of intersection of the horizontal and vertical line serves as the independent variable, with the length of the horizontal and vertical lines being controlled \cite{kunnapas1955analysis,mamassian2010simple,zeyu_li_2017}.

Lastly, the Zollner illusion dataset involves a set of black, uniformly straight lines, manipulated in stimulus configuration. The angle of the stimuli relative to the vertical direction is the independent variable, while the position of the intersection point is controlled \cite{wallace1965measurements,oyama1975determinants,watanabe2011pigeons}.

\subsection{Evaluation Metrics}

For our study, we chose recall as our primary performance metric. Recall, also known as sensitivity or true positive rate, is particularly useful for our context because it focuses on the proportion of actual positive cases that the model correctly identified. In our case, these are instances where the model accurately recognized the in-diagram logic.

The formula for the recall is: 

$$Recall = \frac{True Positive}{True Positive + False Negative}$$

In the context of our study, a true positive is an instance where both the model and the ground truth agree that the sample contains a specific in-diagram logic. A false negative, on the other hand, is a case where the model fails to recognize the in-diagram logic, but it is present according to the ground truth.

The choice of recall as a metric is based on our interest in how accurately the models can detect the presence of in-diagram logic in the samples, regardless of their confidence in the absence of such logic in the negative cases. A high recall score indicates that a model is excellent at detecting in-diagram logic, while a low score could suggest issues with the model's ability to recognize such logic.

By focusing on recall, we aim to ensure that our models are not just accurate on average but are particularly attuned to identifying the nuanced in-diagram logic in our dataset. This makes recall a suitable metric for our goal of understanding the models' ability to interpret in-diagram logic.

\subsection{Baseline Model}

For our benchmark, we choose the Xception model \cite{chollet_2017_xception} as the baseline. Xception, which stands for "Extreme Inception," is an extension of the Inception architecture that replaces the standard Inception modules with depthwise separable convolutions. It is a Convolutional Neural Network (CNN) designed for high-performance image classification tasks.

The Xception model was proposed by François Chollet, the creator of the Keras library, and has been proven to achieve impressive results on several large-scale datasets, including the ImageNet \cite{5206848}. It employs depthwise separable convolutions, which is a form of factorized convolutions that allow the model to use fewer parameters while maintaining a high level of performance. This makes Xception an efficient and powerful model for image classification tasks.

The choice of Xception as the baseline is based on its balanced performance in image classification and in-diagram logic interpretation tasks. It has demonstrated good generalization capabilities across different types of visual content, making it a suitable reference point for evaluating the performance of other models.

\subsection{Ethical Concern}

The application of deep learning models in interpreting in-diagram logic, especially within visual illusions, necessitates an ethical lens. The potential misuse of these models due to insufficient robustness against varying illusion strengths could lead to far-reaching consequences in critical fields. When advancing these models, potential vulnerabilities such as susceptibility to adversarial attacks must be considered to prevent misuse. Therefore, researchers should ensure their models are not only effective but also ethically sound, fair, robust, and secure, continuously addressing ethical issues as the field advances, serving the societies that research ultimately impacts.

\section{Experiment}

\subsection{Experiment Setting}

The experiments were performed on a machine equipped with an RTX3090 GPU. The dataset used for the experiments was composed of 10,000 samples, of which 30\% were positive samples. The remaining 70\% of the samples were negative samples, providing a balanced dataset for the models to learn from.

Ten different models were evaluated in this experiment. These models were chosen to provide a diverse representation of various types of deep learning architectures, including both traditional and more up-to-date models. The performance of the models was evaluated based on their ability to correctly interpret in-diagram logic in the context of visual illusions. 

The models were trained until they reached optimal performance, as determined by a lack of improvement in validation loss over a certain number of epochs. After training, the models were tested on a separate test set to evaluate their generalization performance. The results of these experiments provide insights into the logic interpretation capabilities of the different deep learning models, as well as their strengths and weaknesses in this context.

\subsection{Benchmark Models}

In this study, we evaluate the performance of 10 popular deep-learning models for in-diagram logic interpretation tasks. These models are selected from various classes, including Convolutional Neural Networks (CNNs), Mobile Networks, Inception Networks, Efficient Networks, NAS (Neural Architecture Search) Networks, and ConvNext Networks.

For the training process, we train those models with pre-trained parameters and employ the AdamW optimizer. This approach allows for a fair comparison of the models' performance in in-diagram logic interpretation tasks and provides insights into their respective strengths and weaknesses. The benchmark result is shown in Table~\ref{benchmark_table}

\begin{table}[!ht]
    \centering
    \caption{Benchmark result of 10 models in InDL dataset and ImageNet dataset.}
    \label{benchmark_table}
    \centering
    \resizebox{\linewidth}{!}{
        \begin{tabular}{clcccccccc}
            \toprule
            \multicolumn{2}{c}{} & \multicolumn{6}{c}{InDL recall} & \multicolumn{2}{c}{ImageNet accuracy}                   \\
            \cmidrule(r){3-8} \cmidrule(r){9-10}
            \textbf{year} & \textbf{model} & \textbf{dataset01} & \textbf{dataset02} & \textbf{dataset03} & \textbf{dataset04} & \textbf{dataset05} & \textbf{mean} & \textbf{top 1} & \textbf{top 5} \\ 
            \midrule
            \textbf{2014} & VGG16 \cite{Simonyan2014VeryDC} & 99.49\% & 90.65\% & 85.25\% & 93.41\% & 94.99\% & 92.86\% & 71.59\% & 90.38\%  \\ 
            \textbf{2016} & Inception ResNet V2 \cite{Szegedy2016Inceptionv4IA} & 99.49\% & 87.33\% & 80.65\% & 93.85\% & 89.53\% & 90.27\% & 80.46\% & 95.31\%  \\ 
            \textbf{2017} & Xception \cite{chollet_2017_xception} & 99.49\% & 88.14\% & 83.88\% & 93.85\% & 83.21\% & 89.81\% & 79.05\% & 94.39\%  \\ 
            \textbf{2017} & DenseNet201 \cite{huang2017densely} & 99.49\% & 82.90\% & 84.09\% & 93.85\% & 94.09\% & 90.99\% & 77.29\% & 93.48\%  \\ 
            \textbf{2018} & Darknet53 \cite{Redmon2018YOLOv3AI} & 99.49\% & 83.26\% & 82.23\% & 93.85\% & 83.31\% & 88.53\% & 80.53\% & 95.42\%  \\ 
            \textbf{2018} & NASNetLarge \cite{zoph2018learning} & 99.49\% & 84.44\% & 82.06\% & 93.85\% & 87.25\% & 89.52\% & 82.62\% & 96.05\%  \\ 
            \textbf{2019} & MobileNetV3 \cite{howard2019searching} & 99.49\% & 81.30\% & 74.77\% & 93.85\% & 71.48\% & 84.28\% & 75.77\% & 92.54\%  \\ 
            \textbf{2021} & ResNetV2\_50 \cite{wightman2021resnet} & 82.21\% & 80.81\% & 80.98\% & 93.41\% & 85.22\% & 88.08\% & 80.43\% & 95.08\%  \\ 
            \textbf{2021} & EfficientNetV2 \cite{tan2021efficientnetv2} & 99.49\% & 83.43\% & 70.59\% & 93.85\% & 79.15\% & 85.40\% & 84.81\% & 97.15\%  \\ 
            \textbf{2022} & ConvNext \cite{liu2022convnet} & 99.49\% & 89.42\% & 89.23\% & 93.41\% & 95.30\% & 93.47\% & 87.75\% & 98.55\%  \\ \hline
        \end{tabular}
        }
\end{table}

\subsection{Insights into ImageNet Dataset and InDL Dataset}

In the following analysis, we delve deeper into the performance dichotomy observed across different models, specifically focusing on the performance in the ImageNet \cite{5206848} classification and our InDL classification tasks. This analysis builds upon the preliminary observations shared in the previous sections and offers a comprehensive perspective on the models' responses to varying task complexities and illusion strengths.

\begin{figure}[!ht]
  \centering
  \includegraphics[width=0.765\columnwidth]{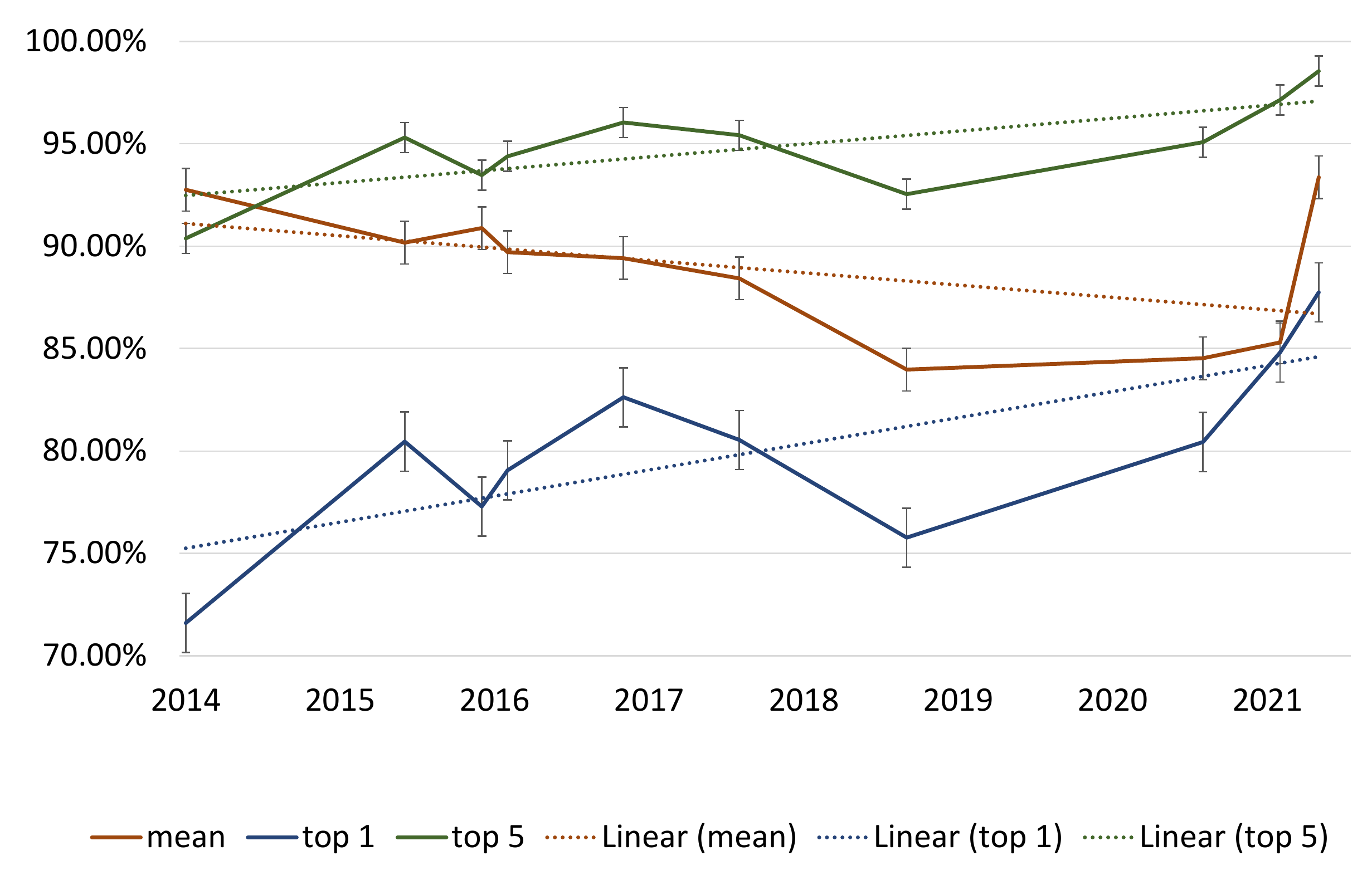}
  \caption{Contrasting trends of performance between ImageNet classification (both top-1 and top-5 accuracy) and in-diagram logic interpretation (mean recall) across various deep learning models.}
  \label{trend-figure}
\end{figure}

As depicted in Figure~\ref{trend-figure}, despite an overall increase in the top-1 and top-5 accuracies on ImageNet over time, we observe a contrasting trend for the recall value on our InDL dataset. This divergence provides additional evidence for the disconnect hypothesized earlier between traditional classification tasks and in-diagram logic interpretation.

The VGG16 model, despite its lower accuracy on ImageNet, excelled on our InDL dataset, whereas the more recent ResNetV2-50 and EfficientNetV2 models, despite their superior ImageNet accuracies, performed comparatively poorer on the InDL dataset. These observations raise critical questions about the evolution of deep learning models, particularly their ability to interpret in-diagram logic across varying task complexities.

\subsection{Unveiling the Inspiration of Illusion on Logic Interpretation with Deep Learning Models}

\begin{figure}[!ht]
  \centering
  \includegraphics[width=\columnwidth]{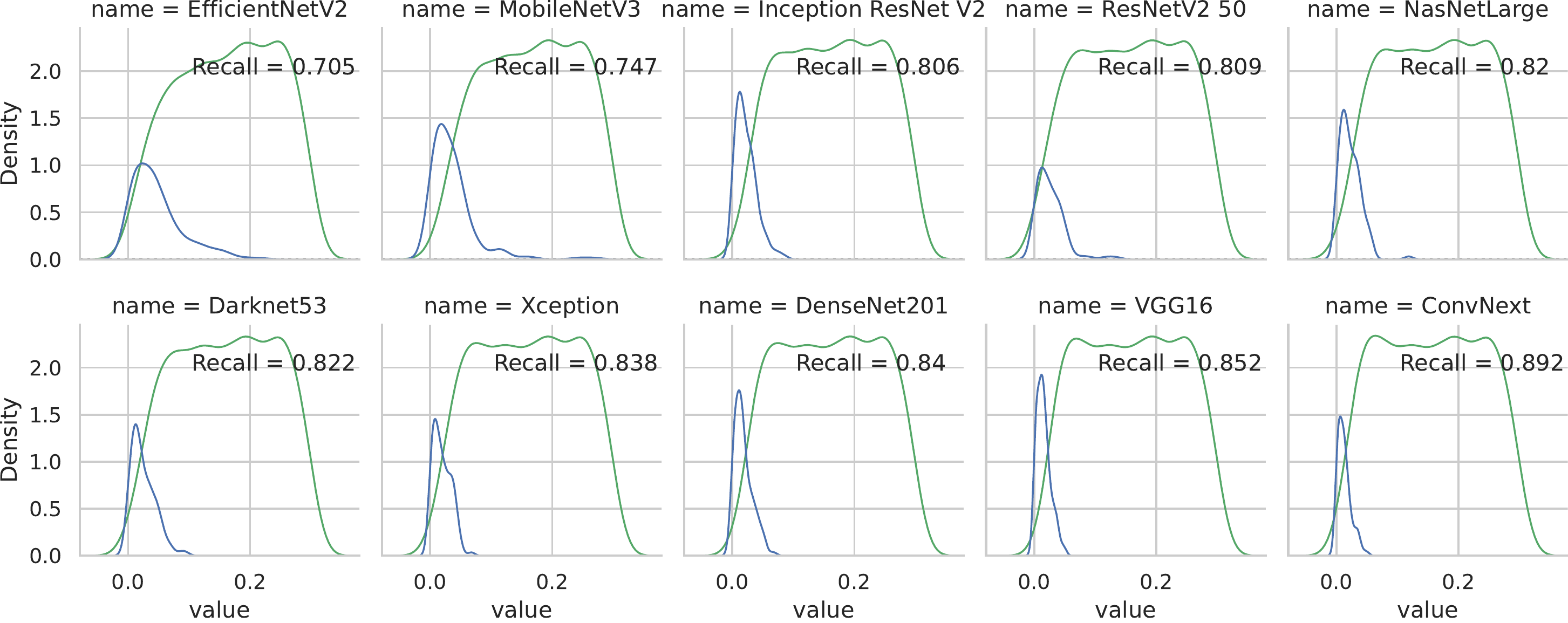}
  \caption{Kernel density estimation (KDE) plot of negative predictions (false negative is the blue line on the left, and true negative is the green line on the right) across various illusion strengths for different deep learning models. The recall values are annotated on each plot.}

  \label{distribution-figure}
\end{figure}

Furthering our analysis, we evaluated the models' responses to the Poggendorff illusion, a prominent component of our InDL dataset, across varying illusion strengths. Our interest was particularly in understanding how the models' recall performance varied with the dispersion of false negative predictions over different illusion strengths, represented in Figure~\ref{distribution-figure}.

Our findings revealed a critical relationship between recall performance and the spread of false negatives. Models demonstrating higher recall performance were observed to maintain consistent performance across varying illusion strengths. However, models with lower recall showcased a greater spread of false negatives, implying a greater susceptibility to the influence of illusion strength variation.

This performance dichotomy could be attributed to the models' ability to interpret the linear relationship inherent in the Poggendorff illusion. As the illusion strength, which corresponds to the angle of the oblique line, varies, the performance of models diverges. A wider range of illusion strength, or in other words, a wider range of oblique angles, seemed to challenge the models' interpretability capacity. Models that excelled in maintaining a consistent performance across this range, such as VGG16, effectively demonstrated their robustness in interpreting linear relationships despite the optical illusion. On the other hand, models like MobileNetV3 and EfficientNetV2, which showed a larger dispersion of false negatives, pointed towards their limitations in comprehending the linear relationship as the illusion strength increased.

These insights underscore the importance of our InDL dataset for exploring the robustness of deep learning models against varying illusion strengths. It also suggests that future research should focus on enhancing models' capabilities to interpret in-diagram logic across varying task complexities, particularly when faced with illusions like the Poggendorff illusion that challenge their interpretation of linear relationships. This is an essential step towards bridging the performance gap we have observed between traditional classification tasks and in-diagram logic interpretation tasks.

\subsection{Influence of Deep Learning Model Depth on Training Results}
\label{sec:depth_influence}

\begin{figure}[!ht]
  \centering
  \subfloat[Training loss and recall curves for different model depths on the InDL dataset.]{
    \includegraphics[width=0.45\textwidth]{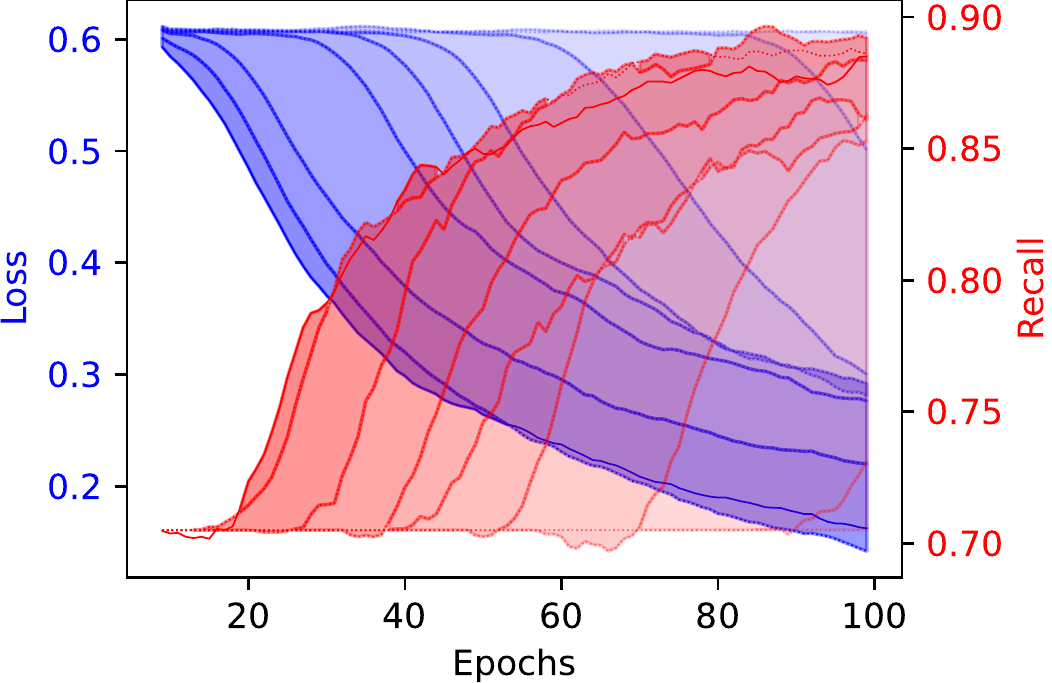}
    \label{fig:loss_recall}
  }
  \quad 
  \subfloat[Training loss and recall curves for different model depths on the MNIST dataset.]{
    \includegraphics[width=0.45\textwidth]{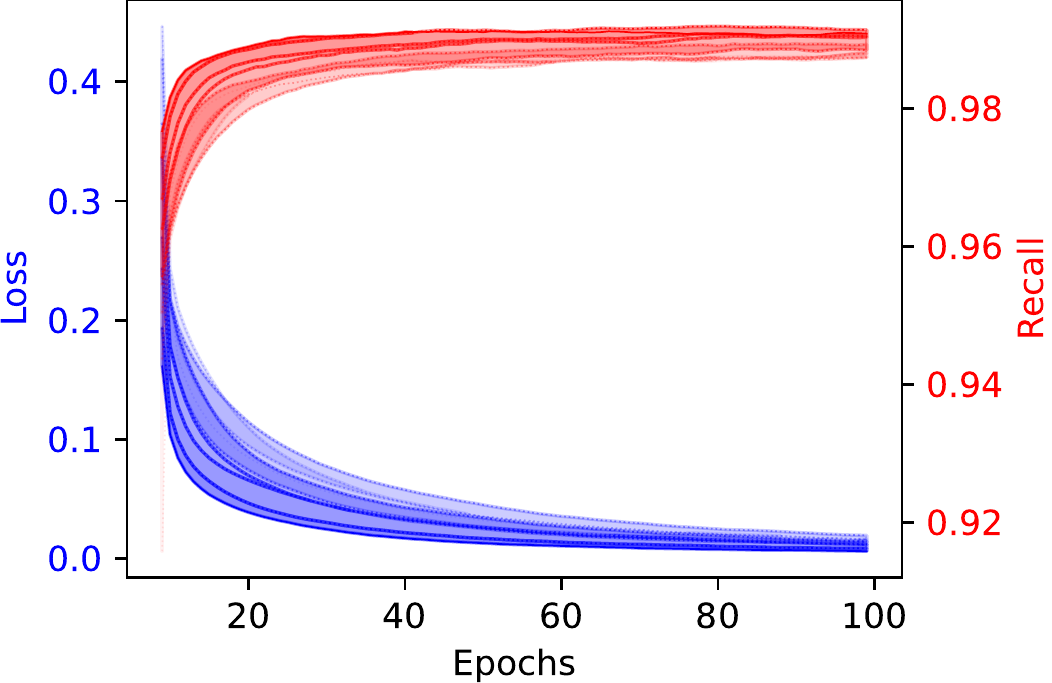}
    \label{fig:mnist_results}
  }
  \caption{Comparison of training loss and recall curves for different model depths on the InDL dataset (a) and the MNIST dataset (b). Each line represents a model of a specific depth, with the opacity of the line indicating the depth of the model (lighter lines represent deeper models). The shaded areas between the lines indicate the difference in performance between consecutive model depths. The contrasting patterns between the two datasets highlight the unique challenges posed by the InDL dataset.}

  \label{fig:double}
\end{figure}

In this experiment, we want to investigate the influence of deep learning model depth on the interpretability of in-diagram logic. Our initial findings, as depicted in Figure~\ref{fig:loss_recall}, suggest that the depth of a model does indeed impact its training results. Specifically, we observed that as the model depth increases, both the recall and loss curves shift to the right, indicating that deeper models find the task of in-diagram logic interpretation more challenging and thus experience a delay in the training process.

To determine whether this curve shift is a universal issue or specific to our task, we conducted the same experiments using the MNIST dataset, which is widely recognized as a representative example for general classification tasks. And the results, shown in Figure~\ref{fig:mnist_results}, did not exhibit the same shift. This contrast suggests that the phenomenon we observed with our InDL dataset is not a general issue across datasets.

Given this finding, it becomes apparent that further research is wanted. For example, a heuristic or a more logic-sensitive method of training may be desirable to accelerate training or even further improve model performance in tasks that requires sensitivity to diagram logic, as discussed in previous sections.

\section{Conclusion and Future Work}

In conclusion, our research offers a fresh perspective into the capabilities of deep learning models, unveiling their strengths and potential weaknesses in interpreting in-diagram logic through the prism of visual illusions. This innovative approach casts light on the opaque nature of these models, potentially catalyzing improvements in their logic interpretation abilities.

Our rigorous quantitative and qualitative analyses, bolstered by the unique InDL dataset, affirm the efficacy of our proposed framework. Intriguing patterns emerged, suggesting a somewhat paradoxical relationship between a model's proficiency in handling the ImageNet dataset and its performance on InDL datasets. This insight underlines the importance of targeted benchmarking frameworks to truly understand and optimize deep learning models for specific tasks.

In the realm of future work, several promising paths lie ahead. One compelling extension of our research would be to intensify the complexity of the visual illusions and logic scenarios in our dataset, pushing the boundaries of current deep learning models. Moreover, our work hints at a rich seam of exploration where psychological phenomena observed in humans are emulated in deep learning models, suggesting that this research approach could unlock further insights into model comprehension and performance. Furthermore, we anticipate that our evaluation methodology could be adapted and applied to other domains, such as natural language processing or reinforcement learning, thereby amplifying its reach and impact.

\printbibliography

\end{document}